\documentclass[10pt,twocolumn,letterpaper]{article}

\usepackage{iccv}
\usepackage{times}
\usepackage{epsfig}
\usepackage{graphicx}
\usepackage{amsmath}
\usepackage{amssymb}
\usepackage[table]{xcolor}
\usepackage{algorithm}
\usepackage[noend]{algpseudocode}
\usepackage{enumitem}
\usepackage{url}
\usepackage{endnotes}
\let\footnote=\endnote

\usepackage[pagebackref=true,breaklinks=true,letterpaper=true,colorlinks,bookmarks=false]{hyperref}

\iccvfinalcopy 

\def\iccvPaperID{9} 

\ificcvfinal\fi
\begin{document}

\title{Dual Path Networks for Multi-Person Human Pose Estimation}

\author{Guanghan Ning, Zhihai He \\
	University of Missouri\\
	Columbia, MO\\
	{\tt\small gnxr9@mail.missouri.edu, hezhi@missouri.edu}
}		

\maketitle

\begin{abstract}
The task of multi-person human pose estimation in natural scenes is quite challenging. Existing methods include both top-down and bottom-up approaches. The main advantage of bottom-up methods is its excellent tradeoff between estimation accuracy and computational cost. We follow this path and aim to design smaller, faster, and more accurate neural networks for the regression of keypoints and limb association vectors. These two regression tasks are naturally dependent on each other. In this work, we propose a dual-path network\cite{chen2017dual} specially designed for multi-person human pose estimation, and compare our performance with the openpose\cite{openpose, cao2016realtime} network in aspects of model size, forward speed, and estimation accuracy.
\end{abstract}

\section{Introduction}

The task of human pose estimation is to determine the precise pixel locations of body keypoints from a single input image \cite{insafutdinov2016deepercut, newell2016stacked}.
Human pose estimation is very important for many high-level computer vision tasks, including action and activity recognition, human-computer interaction, motion capture, and animation.
Estimating human poses from natural images is quite challenging. An effective pose estimation system must be able to handle large pose variations, changes in clothing and lighting conditions, severe body deformations, heavy body occlusions \cite{toshev2014deeppose, tompson2014joint, newell2016stacked}.
It is naturally a regression task. 
With Convolutional Neural Networks (ConvNets) and many assistive methods such as batch normalization \cite{ioffe2015batch}, resnet \cite{he2016deep}, and inception design \cite{szegedy2015going, szegedy2016rethinking}, single-person human pose estimation has recently achieved significant progress.

Recent research emphasis has been put on multi-person human pose estimation, where multiple individuals may exist in a natural scene. Compared to single person human pose estimation, where human candidates are cropped and centered in the image patch, the task of multi-person human pose estimation is more challenging. The best performance on MS COCO 2016 Keypoints challenge \cite{cocokeypoint2016} is only around 60\% in mean average precision (mAP).

Existing methods can be classified into two kinds of approaches, the top-down approach and the bottom-up approach. The top-down approach \cite{fang2017rmpe, papandreou2017towards} relies on a detection module to obtain human candidates and then apply a single-person human pose estimator to detect human keypoints. The bottom-up approach \cite{cao2016realtime, he2017mask, xia2017joint, newell2016associative}, on the other hand, detects human keypoints from all potential human candidates and then assemble these keypoints into human limbs for each individual based on various data association techniques. 
The main advantage of the bottom-up approaches is its excellent tradeoff between estimation accuracy and computational cost. It takes the winner \cite{cao2016realtime} of MS COCO 2016 keypoint challenge less than 200 ms to run the pose estimator for one frame on a Pascal TITAN X GPU. More importantly, contrary to top-down approaches, its computational cost is invariant to the number of human candidates in the image.
We follow the bottom-up approach of the works of Zhe. et. al \cite{cao2016realtime} and aim to design smaller, faster, and more accurate neural networks for multi-person keypoints regression. According to \cite{cao2016realtime}, their proposed Part Affinity Fields (PAF) and its corresponding data-association techniques are robust and reliable. More accurate keypoint and PAF regression would potentially increase the overall performance up to 10\%.

In this work, we focus on the network regression part and leave the data association part to future works. We propose a dual-path network specially designed for multi-person human pose estimation, and compare our performance with the openpose \cite{openpose} network in aspects of model size, forward speed, and estimation accuracy.
Our contributions include: (1) We analyze the tasks of keypoint regression and PAF, where PAF estimation depends heavily on keypoints estimation but not vice versa. (2) We then design a dual-path network, the denseNet path responsible for PAF regression while the resNeXt path regressing human keypoints. Our performance is superior than the openpose \cite{openpose} network even though the proposed network is of lower computational complexity and smaller model size.    

\begin{figure*}[t]
	\begin{center}
		\includegraphics[width=1.0\linewidth]{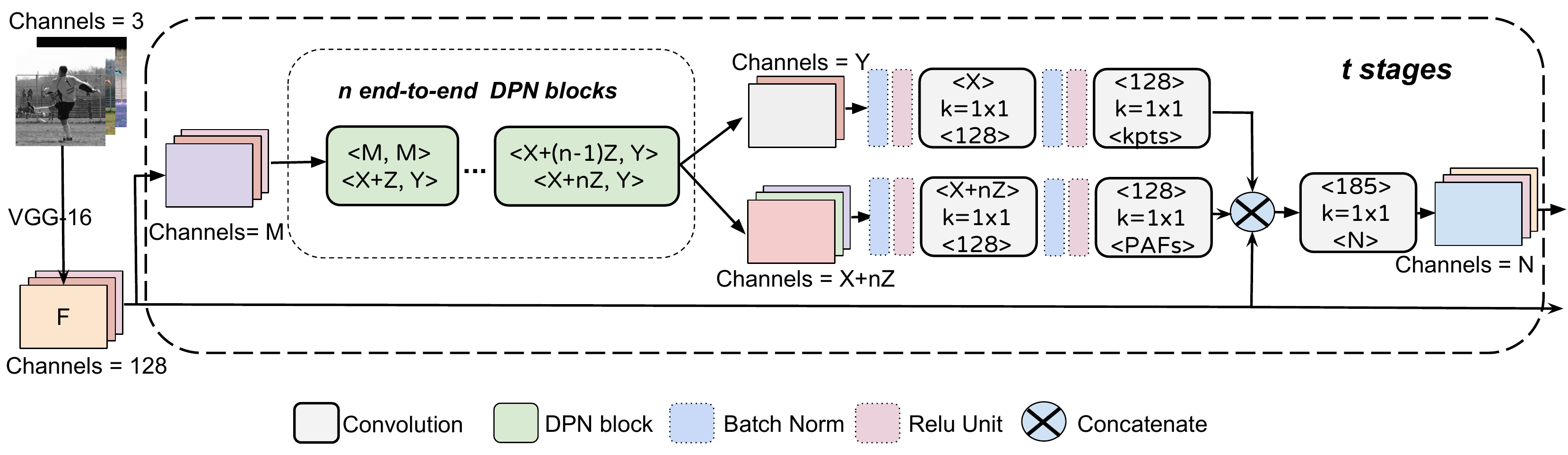}
	\end{center}
	\caption{\textbf{Our proposed network}. The image feeds into the first 16 layers of VGG \cite{simonyan2014very} network and outputs 128 channels of low-level visual features. These features are subsequently fed into each stage of the repetive sub-network illustrated in this image. Each DPN block takes as input two branches of feature maps and also output feature maps of two branches, one with a consistent number of channels, the other one with accumulated channels over DPN blocks.}
	\label{fig:network}
\end{figure*}

The rest of the paper is organized as follows.
In section \ref{sec:related-work}, we provide a brief review of recent works on multi-person human pose estimation. Section \ref{sec:proposed-method} introduces the proposed network. Section \ref{sec:experiments} presents our experimental results. Section \ref{sec:conclusions} concludes our paper.

\section{Related Work}
\label{sec:related-work}

\subsection{Single-Person Human Pose Estimation}
This task is simpler than multi-person pose estimation because it aims to estimate the pose of a single person, where the image is cropped assuming the person dominates the image content.
Traditional methods for single-person human pose estimation are mostly based on pictorial structure models \cite{sapp2013modec, pishchulin2013poselet, sun2011articulated, tian2012exploring, dantone2013human, karlinsky2012using}. 
Since the work of \textit{DeepPose} by Toshev {\it et al.} \cite{toshev2014deeppose}, research on human pose estimation has shifted from traditional approaches to deep neural networks (DNN) due to their superior performance. 
Recent methods \cite{newell2016stacked, wei2016convolutional, insafutdinov2016deepercut, belagiannis2016recurrent} have achieved quite accurate performance on popular datasets \cite{andriluka20142d, johnson2010clustered}.  
However, the assumption that the person can always be correctly located is not necessarily satisfied.

\subsection{Multi-Person Human Pose Estimation}
Multi-person human pose estimation is a more realistic problem. It attempts to estimate the poses of multiple persons in natural scenes. It is quite challenging due to the variance of sizes and scales of the persons.  
Existing methods can be classified into two kinds of approaches, the top-down approach and the bottom-up approach. 

The top-down approach \cite{fang2017rmpe, papandreou2017towards} relies on a detection module to obtain human candidates and then apply a single-person human pose estimator to detect human keypoints. 
Insafutdinov \textit{et al} \cite{insafutdinov2016deepercut} propose a pipeline which uses the Faster R-CNN \cite{ren2015faster} as detection module and a unary DeeperCut as their single-person pose estimator. Their method achieves 51.0 in mAP on MPII dataset \cite{andriluka20142d}. 
Because the single-person pose estimator is usually sensitive to the detection results, this approach requires the detection module to be very robust. 
More accurate performance has been achieved by Hao \textit{et al} \cite{fang2017rmpe}. Their framework facilitates pose estimation in the presence of inaccurate human bounding boxes by introducing more components into the pipeline that refine the detection and pose estimation results.

The bottom-up approach \cite{cao2016realtime, he2017mask, xia2017joint, newell2016associative}, on the other hand, detects human keypoints from all potential human candidates and then assemble these keypoints into human limbs for each individual based on various data association techniques. Many of these techniques are graph-based \cite{cao2016realtime, xia2017joint}. 
The main advantage of the bottom-up approaches is its excellent tradeoff between estimation accuracy and computational cost. 
The winner of COCO2016 \cite{cocokeypoint2016} proposes to estimate human keypoints as well as \textit{Part-Affinity Fields} (PAF) simultaneously. PAFs are limb association vectors that can be used to assemble the keypoints into multi-person poses with certain graph-based association techniques. 
According to \cite{cao2016realtime}, their proposed PAF and corresponding data-association techniques are robust and reliable. More accurate keypoint and PAF regression would potentially increase the overall performance up to 10\%.
We follow their works and focus on the network regression part, aiming to design smaller, faster, and more accurate neural networks for multi-person human pose estimation.

\subsection{Dual Path Networks}

\begin{figure*}[t]
	\begin{center}
		\includegraphics[width=0.9\linewidth]{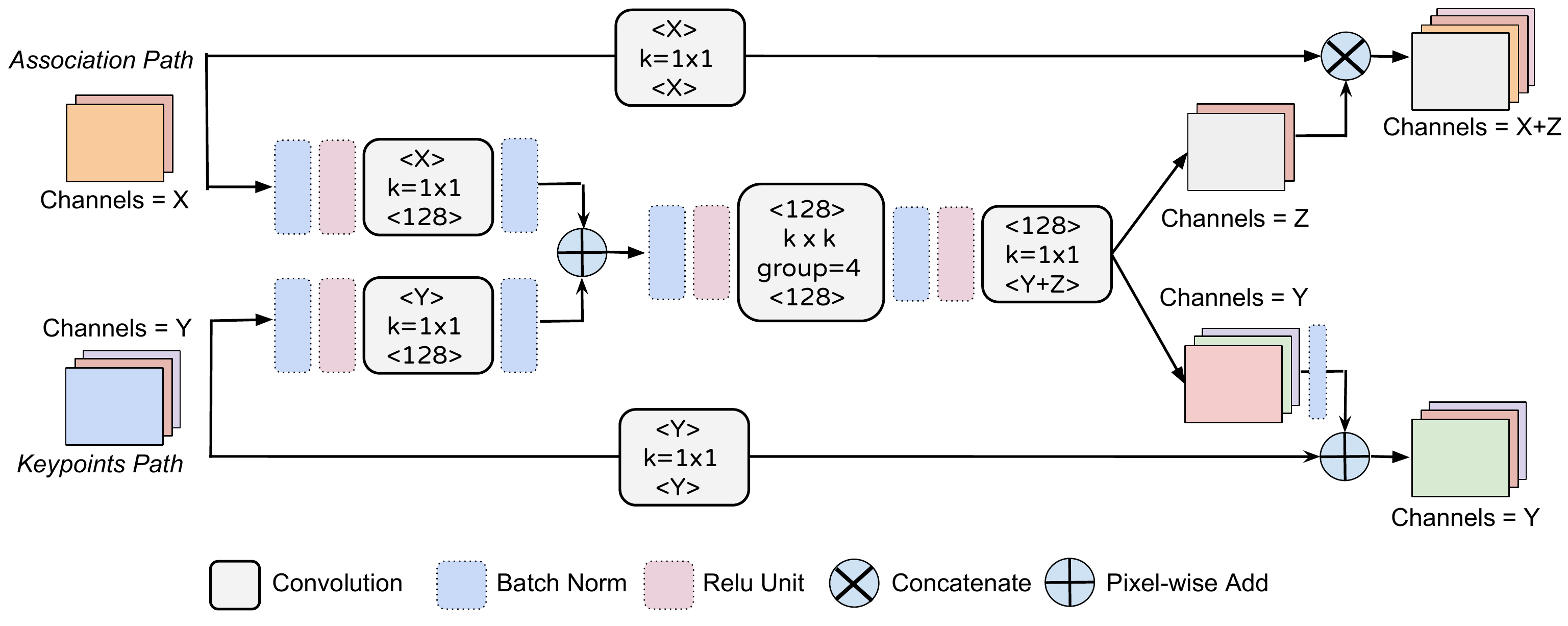}
	\end{center}
	\caption{\textbf{Proposed DPN block}. It consists of two paths: the keypoints path and the data association path. The regression of human keypoints and association vectors are dependent on each other and share information from the previous block. The association vectors however, need more features to further explore spatial interdependency and they are regressed with accumulated channels of feature maps from all previous DPN blocks.}
	\label{fig:DPN_block}
\end{figure*}

According to \cite{cao2016realtime}, their proposed data-association technique is robust and reliable; more accurate keypoint and PAF regression would potentially increase the overall performance up to 10\%. Motivated by this, we look into network engineering and explore more robust and efficient learning of features and spatial inter-dependencies.

Dual Path Networks (DPN) is first proposed in \cite{chen2017dual} as a hybrid network design that incorporates the core idea of DenseNet \cite{huang2016densely} with that of ResNeXt \cite{xie2016aggregated}. 
ResNeXt is a variant of the widely-used ResNet \cite{he2016deep}, introducing a homogeneous, multi-branch architecture that has a new dimension called \textit{cardinality} (the size of the set of transformations), as an essential factor in addition to the dimensions of depth and width.
They show that increasing cardinality is able to improve classification accuracy, and is more effective than going deeper or wider when we increase the capacity of the network.
The core of denseNet is that it connects each layer to every other layer in a feed-forward fashion. They alleviate the vanishing-gradient problem, strengthen feature propagation, encourage feature reuse, and substantially reduce the number of parameters.

According to the research of DPN, ResNet and its variants enable feature re-usage while DenseNet enables new features exploration which are both important for learning good representations.
By carefully incorporating these two network designs into dual-path topologies, DPN shares common features while maintaining the flexibility to explore new features through dual path architectures.

Inspired by the DPN network that is originally designed for the task of image classfication, we aim to design a variant of DPN that is specially tailored for multi-person human pose estimation because the regression tasks for keypoints and association vectors are naturally two paths. The two tasks are dependent on each other but unique in their own ways.
In the next section, we introduce our proposed DPN network, describe how the regression of keypoints and association vectors are assigned to each path and,  explain the intuition behind it. 
For detailed description of part association techniques, please refer to the original paper of PAF \cite{cao2016realtime}.

\section{Proposed Method}
\label{sec:proposed-method}

The proposed network is highly modulized. We intentionally follow the general network structure of openpose. As shown in Figure \ref{fig:network}, there are multiple stages of repetitive subnetworks, where each subnetwork outputs estimated heatmaps of keypoints and PAFs and is enforced with loss functions as intermediate supervision. The network is first fed with an image into the first 16 layers of the VGG network and outputs 128 channels of low-level visual features, denoted by $\mathbf{F}$. These features are then fed into each stage of the following subnetworks. The modules in the figure only indicates the input and output channels and leaves out the resolution because the convolutional layers are all padded such that the resolution of the feature maps do not change.  

Our proposed network differs from the openpose network in the structure of the subnetwork patterns, specifically, the DPN blocks.
As shown in Figure \ref{fig:DPN_block}, our proposed DPN block consists of two paths. The regression of human keypoints and association vectors are dependent on each other and share information from the previous block. 
With the operator of element-wise addition, the \textit{Keypoints Path} (KP) leverages features before and after the feature fusion/transition within a DPN block.  
The association vectors in the \textit{Association Path} (AP), however, accumulate features over blocks to further exploit spatial interdependencies. They are regressed with accumulated channels of feature maps from all previous DPN blocks.
With such representation, features from the AP path is less constrained and more flexible than the KP path. It enforces the AP path to learn features at a higher level compared to the KP path, even though they are dependent and share common features.

It is declared in \cite{cao2016realtime} that most of their false positives come from imprecise localization, other than background confusion and that there is more improvement space in capturing spatial dependencies than in recognizing body parts appearances. Therefore, we set the learning rate for the VGG layers to be zero, thus maintaining the same low-level visual features as that used in the openpose model. In this way, we can purely compare the capability of the networks in capturing spatial dependencies. 

The network from the first stage produces a set of keypoint heatmaps $\mathbf{S}^1 = \rho^{1}(\mathbf{F})$ and a set of PAFs $\mathbf{L}^1 = \phi^{1}(\mathbf{F})$, where $\rho^1$ and $\phi^1$ represent high-dimensional functions of the KP path and AP path networks.
To guide the network to iteratively predict keypoint heatmaps and PAFs at each stage, we apply two loss functions in each sub-network. We use an $L_2$ loss between the estimated predictions and the groundtruth maps and fields. Specifically, the loss functions for the dual paths at stage $t$ are:
 \begin{eqnarray}
f^t_\mathbf{S} &=& \sum_{j = 1}^{J} \sum_{\mathbf{p}}   \| \mathbf{S}^t_{j}(\mathbf{p})  -  \mathbf{S}_{j}^{*}(\mathbf{p}) \|^{2}_{2},\\
f^t_\mathbf{L} &=& \sum_{c = 1}^{C} \sum_{\mathbf{p}}  \| \mathbf{L}^t_{c}(\mathbf{p})  -  \mathbf{L}_{c}^{*}(\mathbf{p}) \|^{2}_{2},
\label{eqn:localobjective2}
\end{eqnarray}
where $\mathbf{S}_{j}^{*}$ is the groundtruth keypoint heatmap, $\mathbf{L}_{c}^{*}$ is the groundtruth PAF vector field,
at an image location $\mathbf{p}$.

\section{Experimental Results}
\label{sec:experiments}

\textbf{Dataset}
The PoseTrack \cite{poseTrack} dataset consists of over $68,000$ frames.
The workshop is organized around a challenge with three competition tracks focusing on single frame multi-person pose estimation, multi-person pose estimation in videos, and multi-person articulated tracking. In our work, we focus on the single frame multi-person pose estimation.

\textbf{Experimental Settings}
In order to make a fair comparison with the openpose network, which is trained on the MS COCO dataset \cite{cocokeypoint2016}, we use the same training data before testing on the PoseTrack dataset. In our experiments, all the experiment settings including the testing scales and parameters in the data association techniques are uniform for the two networks. Therefore, no special tuning on the training and testing for the PoseTrack dataset is made.

\textbf{Quantitative Results}
We report our Average Precision (AP) scores on the PoseTrack test set\footnote{Performance on test set is evalutaed by the PoseTrack server. Challenge results available at: \url{https://posetrack.net/workshops/iccv2017/posetrack-challenge-results.html}}. The result on test set is performed at 3 stages and 2 scales (1, 0.75).

\begin{table}[H]
	\centering 
	\small
	\tabcolsep=0.07cm
	\begin{tabular}{l c c c c c c c c }
		\hline 
		Method & Head & Sho. & Elb. & Wri. & Hip & Knee & Ank. & Total\\ [0.1ex]
		\hline 
		Ours  & 48.2 & 75.4 & 68.8 & 59.5 & 63.6 & 60.1 & 53.9 & 62.4 \\
		\hline
		
	\end{tabular}
	\caption{Average Precision (AP) scores on the PoseTrack test set. }
	\label{table:poseTrack-test}
\end{table}

\subsection{Algorithm Performance Analysis}

\textbf{Average Precision}
We compare the AP scores of the proposed network with openpose on the validation set. All experiments are performed on the local machine with the same resolution and single scale.
\begin{table}[H]
	\centering 
	\small
	\tabcolsep=0.07cm
	\begin{tabular}{l c c c c c c c c }
		\hline 
		Method & Head & Sho. & Elb. & Wri. & Hip & Knee & Ank. & Total\\ [0.1ex]
		\hline 
		openpose@6stages & 46.8 & 76.4 & 68.7 & 54.7 & 63.6 & 59.6 & 52.8 & 59.5 \\
		openpose@3stages  & 45.5 & 72.1 & 63.1 & 48.1 & 58.4 & 51.9 & 45.8 & 54.4 \\
		\hline	
		Ours@3stages & 47.5 & 76.3 & 67.6 & 53.3 & 62.9 & 57.9 & 49.7 & 58.5 \\
		\hline
		
	\end{tabular}
	\caption{Comparisons of Average Precision (AP) scores on the PoseTrack validation set. Experiments are performed at the same original resolution and single scale.}
	\label{table:poseTrack-val}
\end{table}

\textbf{Speed and Model Size Comparison}
We test and compare the networks by averaged forward time for a single frame. The unit is miliseconds (ms). Both experiments are performed at the same original resolution and single scale.

\begin{table}[H]
	\centering 
	\small
	\tabcolsep=0.07cm
	\begin{tabular}{l c c }
		\hline 
		Method & forward time (ms) \\ 
		\hline 
		openpose@6stages & 155.8 \\ 
		Ours@3stages & 186.6 \\ 
		\hline
		
	\end{tabular}
	\caption{Comparisons of feedforward time in miliseconds (ms) of different networks for a single frame. Evaluations are performed with a single Pascal TITAN X GPU.}
	\label{table:poseTrack-time}
\end{table}

\begin{table}[H]
	\centering 
	\small
	\tabcolsep=0.07cm
	\begin{tabular}{l c c c c c c c c }
		\hline 
		Method & 3 stages & 4 stages & 5 stages & 6 stages \\ [0.1ex]
		\hline 
		openpose & 103.8 & 139.0 & 174.1 & 209.3  \\
		Ours & 43.7 & 50.1 & 56.4 & 62.7  \\
		\hline
		
	\end{tabular}
	\caption{Comparisons of model size of different networks in Mega Bytes (MB). Both models have the same input image resolution and share the same VGG-16 layers.}
	\label{table:model-size}
\end{table}

Even though our model is much smaller than the openpose model, the intermediate storage of network strucures including accumulated feature maps from AP path consume GPU memory greatly in current Caffe \cite{jia2014caffe} version. 
In the future, by porting memory-efficient denseNet implementation from other deep learning frameworks \cite{amos2017pytorch, abadi2016tensorflow} into Caffe, which enables more stages of DPN to fit into the GPU memory, we believe the performance will potentially be better.

\section{Conclusion}
\label{sec:conclusions}
In this work, we propose a dual-path network specially designed for multi-person human pose estimation, and compare our performance with the openpose\cite{cao2016realtime} network in aspects of model size, forward speed, and estimation accuracy. Extentive experiments on PoseTrack challenge dataset 
show that our method is both accurate and efficient. 
Even though the method described in this work regresses PAFs\cite{cao2016realtime} as the association vector, the dual-path network is generic and not limited to specific vector representation and association techniques. 
{\small
\bibliographystyle{ieee}
\bibliography{ning}

\begin{thebibliography}{10}\itemsep=-1pt

\bibitem{cocokeypoint2016}
Coco keypoints challenge.
\newblock \url{http://image-net.org/challenges/ilsvrc+coco2016}, 2016.

\bibitem{openpose}
Openpose library.
\newblock May, 2017.

\bibitem{poseTrack}
Posetrack: Iccv workshop.
\newblock \url{https://posetrack.net/workshops/iccv2017/}, October, 2017.

\bibitem{abadi2016tensorflow}
M.~Abadi, A.~Agarwal, P.~Barham, E.~Brevdo, Z.~Chen, C.~Citro, G.~S. Corrado,
  A.~Davis, J.~Dean, M.~Devin, et~al.
\newblock Tensorflow: Large-scale machine learning on heterogeneous distributed
  systems.
\newblock {\em arXiv preprint arXiv:1603.04467}, 2016.

\bibitem{amos2017pytorch}
B.~Amos and J.~Z. Kolter.
\newblock A pytorch implementation of densenet, 2017.

\bibitem{andriluka20142d}
M.~Andriluka, L.~Pishchulin, P.~Gehler, and B.~Schiele.
\newblock 2d human pose estimation: New benchmark and state of the art
  analysis.
\newblock In {\em CVPR}, 2014.

\bibitem{belagiannis2016recurrent}
V.~Belagiannis and A.~Zisserman.
\newblock Recurrent human pose estimation.
\newblock {\em arXiv preprint arXiv:1605.02914}, 2016.

\bibitem{cao2016realtime}
Z.~Cao, T.~Simon, S.-E. Wei, and Y.~Sheikh.
\newblock Realtime multi-person 2d pose estimation using part affinity fields.
\newblock {\em arXiv preprint arXiv:1611.08050}, 2016.

\bibitem{chen2017dual}
Y.~Chen, J.~Li, H.~Xiao, X.~Jin, S.~Yan, and J.~Feng.
\newblock Dual path networks.
\newblock {\em arXiv preprint arXiv:1707.01629}, 2017.

\bibitem{dantone2013human}
M.~Dantone, J.~Gall, C.~Leistner, and L.~Van~Gool.
\newblock Human pose estimation using body parts dependent joint regressors.
\newblock In {\em CVPR}, 2013.

\bibitem{fang2017rmpe}
Y.-W.~T. Hao-Shu~Fang, Shuqin~Xie and C.~Lu.
\newblock {RMPE}: Regional multi-person pose estimation.
\newblock In {\em ICCV}, 2017.

\bibitem{he2017mask}
K.~He, G.~Gkioxari, P.~Doll{\'a}r, and R.~Girshick.
\newblock Mask r-cnn.
\newblock {\em arXiv preprint arXiv:1703.06870}, 2017.

\bibitem{he2016deep}
K.~He, X.~Zhang, S.~Ren, and J.~Sun.
\newblock Deep residual learning for image recognition.
\newblock In {\em Proceedings of the IEEE conference on computer vision and
  pattern recognition}, pages 770--778, 2016.

\bibitem{huang2016densely}
G.~Huang, Z.~Liu, K.~Q. Weinberger, and L.~van~der Maaten.
\newblock Densely connected convolutional networks.
\newblock {\em arXiv preprint arXiv:1608.06993}, 2016.

\bibitem{insafutdinov2016deepercut}
E.~Insafutdinov, L.~Pishchulin, B.~Andres, M.~Andriluka, and B.~Schiele.
\newblock Deepercut: A deeper, stronger, and faster multi-person pose
  estimation model.
\newblock In {\em ECCV}, 2016.

\bibitem{ioffe2015batch}
S.~Ioffe and C.~Szegedy.
\newblock Batch normalization: Accelerating deep network training by reducing
  internal covariate shift.
\newblock {\em arXiv preprint arXiv:1502.03167}, 2015.

\bibitem{jia2014caffe}
Y.~Jia, E.~Shelhamer, J.~Donahue, S.~Karayev, J.~Long, R.~Girshick,
  S.~Guadarrama, and T.~Darrell.
\newblock Caffe: Convolutional architecture for fast feature embedding.
\newblock In {\em Proceedings of the 22nd ACM international conference on
  Multimedia}, pages 675--678. ACM, 2014.

\bibitem{johnson2010clustered}
S.~Johnson and M.~Everingham.
\newblock Clustered pose and nonlinear appearance models for human pose
  estimation.
\newblock In {\em BMVC}, 2010.

\bibitem{karlinsky2012using}
L.~Karlinsky and S.~Ullman.
\newblock Using linking features in learning non-parametric part models.
\newblock In {\em ECCV}, 2012.

\bibitem{newell2016associative}
A.~Newell and J.~Deng.
\newblock Associative embedding: End-to-end learning for joint detection and
  grouping.
\newblock {\em arXiv preprint arXiv:1611.05424}, 2016.

\bibitem{newell2016stacked}
A.~Newell, K.~Yang, and J.~Deng.
\newblock Stacked hourglass networks for human pose estimation.
\newblock In {\em ECCV}, 2016.

\bibitem{papandreou2017towards}
G.~Papandreou, T.~Zhu, N.~Kanazawa, A.~Toshev, J.~Tompson, C.~Bregler, and
  K.~Murphy.
\newblock Towards accurate multi-person pose estimation in the wild.
\newblock {\em arXiv preprint arXiv:1701.01779}, 2017.

\bibitem{pishchulin2013poselet}
L.~Pishchulin, M.~Andriluka, P.~Gehler, and B.~Schiele.
\newblock Poselet conditioned pictorial structures.
\newblock In {\em CVPR}, 2013.

\bibitem{ren2015faster}
S.~Ren, K.~He, R.~Girshick, and J.~Sun.
\newblock Faster r-cnn: Towards real-time object detection with region proposal
  networks.
\newblock In {\em Advances in neural information processing systems}, pages
  91--99, 2015.

\bibitem{sapp2013modec}
B.~Sapp and B.~Taskar.
\newblock Modec: Multimodal decomposable models for human pose estimation.
\newblock In {\em CVPR}, 2013.

\bibitem{simonyan2014very}
K.~Simonyan and A.~Zisserman.
\newblock Very deep convolutional networks for large-scale image recognition.
\newblock {\em arXiv preprint arXiv:1409.1556}, 2014.

\bibitem{sun2011articulated}
M.~Sun and S.~Savarese.
\newblock Articulated part-based model for joint object detection and pose
  estimation.
\newblock In {\em ICCV}, 2011.

\bibitem{szegedy2015going}
C.~Szegedy, W.~Liu, Y.~Jia, P.~Sermanet, S.~Reed, D.~Anguelov, D.~Erhan,
  V.~Vanhoucke, and A.~Rabinovich.
\newblock Going deeper with convolutions.
\newblock In {\em CVPR}, 2015.

\bibitem{szegedy2016rethinking}
C.~Szegedy, V.~Vanhoucke, S.~Ioffe, J.~Shlens, and Z.~Wojna.
\newblock Rethinking the inception architecture for computer vision.
\newblock In {\em CVPR}, 2016.

\bibitem{tian2012exploring}
Y.~Tian, C.~L. Zitnick, and S.~G. Narasimhan.
\newblock Exploring the spatial hierarchy of mixture models for human pose
  estimation.
\newblock In {\em ECCV}, 2012.

\bibitem{tompson2014joint}
J.~J. Tompson, A.~Jain, Y.~LeCun, and C.~Bregler.
\newblock Joint training of a convolutional network and a graphical model for
  human pose estimation.
\newblock In {\em NIPS}, 2014.

\bibitem{toshev2014deeppose}
A.~Toshev and C.~Szegedy.
\newblock Deeppose: Human pose estimation via deep neural networks.
\newblock In {\em CVPR}, 2014.

\bibitem{wei2016convolutional}
S.-E. Wei, V.~Ramakrishna, T.~Kanade, and Y.~Sheikh.
\newblock Convolutional pose machines.
\newblock In {\em CVPR}, 2016.

\bibitem{xia2017joint}
F.~Xia, P.~Wang, X.~Chen, and A.~Yuille.
\newblock Joint multi-person pose estimation and semantic part segmentation.
\newblock {\em arXiv preprint arXiv:1708.03383}, 2017.

\bibitem{xie2016aggregated}
S.~Xie, R.~Girshick, P.~Doll{\'a}r, Z.~Tu, and K.~He.
\newblock Aggregated residual transformations for deep neural networks.
\newblock {\em arXiv preprint arXiv:1611.05431}, 2016.

\end{thebibliography}
}

\theendnotes
\end{document}